\documentclass[]{fairmeta}

\usepackage{booktabs}
\usepackage{url}
\usepackage{hyperref}
\usepackage{graphicx}
\usepackage{amsmath}
\usepackage{amssymb}

\usepackage{xcolor}
\usepackage{pifont}
\usepackage{gensymb}
\usepackage{colortbl}
\usepackage{multirow}
\usepackage{tabularx}
\usepackage{textcomp}
\usepackage{rotating}
\usepackage{adjustbox}
\usepackage{subcaption}
\usepackage{algpseudocode}
\usepackage[table]{xcolor}

\newcommand{\methodname}{VLM$^{3}$}

\title{\methodname: Vision Language Models Are Native 3D Learners}

\author[1,\dagger]{Zhipeng Cai}
\author[2]{Zhuang Liu}
\author[1]{Yunyang Xiong}
\author[1]{Zechun Liu}
\author[1]{Vikas Chandra}
\author[1]{Yangyang Shi}

\affiliation[1]{Meta}
\affiliation[2]{Princeton University}

\contribution[\dagger]{Project Lead}

\abstract{Vision Language Models (VLMs) enable a unified model to solve various vision tasks through prompting. They have shown promising performance in semantic understanding. However, 3D understanding still largely relies on expert vision models with complex task-specific designs. The key argument this work wants to make is that \emph{VLMs are native 3D learners}. Our in-depth large scale study shows that 1) focal length unification, 2) text-based pixel reference and 3) data mixture and scaling, are all you need for effective 3D learning. Model architecture changes, large models, heavy data augmentations, and complex losses including the regression formulation, many of which form the foundation of expert vision models, are actually \emph{not} necessary conditions. As a result, we propose \emph{\methodname}, a scalable method with the simplest design that enables standard VLMs to master diverse 3D tasks. \emph{\methodname} not only advances the VLM depth estimation accuracy by a large margin (0.84 $\rightarrow$ 0.9), but also enables diverse 3D tasks such as pixel correspondence, camera pose estimation and object-level 3D understanding, matching expert vision model accuracy while maintaining standard architectures and text-based training. We believe \methodname\ opens up a new paradigm for simple and scalable 3D learning.}

\date{\today}
\correspondence{Zhipeng Cai \email{czptc2h@gmail.com}}

\metadata[Code]{\url{https://github.com/facebookresearch/VLM3}}

\begin{document}

\maketitle

\section{Introduction}\label{sec:intro}

\begin{figure}[!htb]
    \centering
    \includegraphics[width=\linewidth]{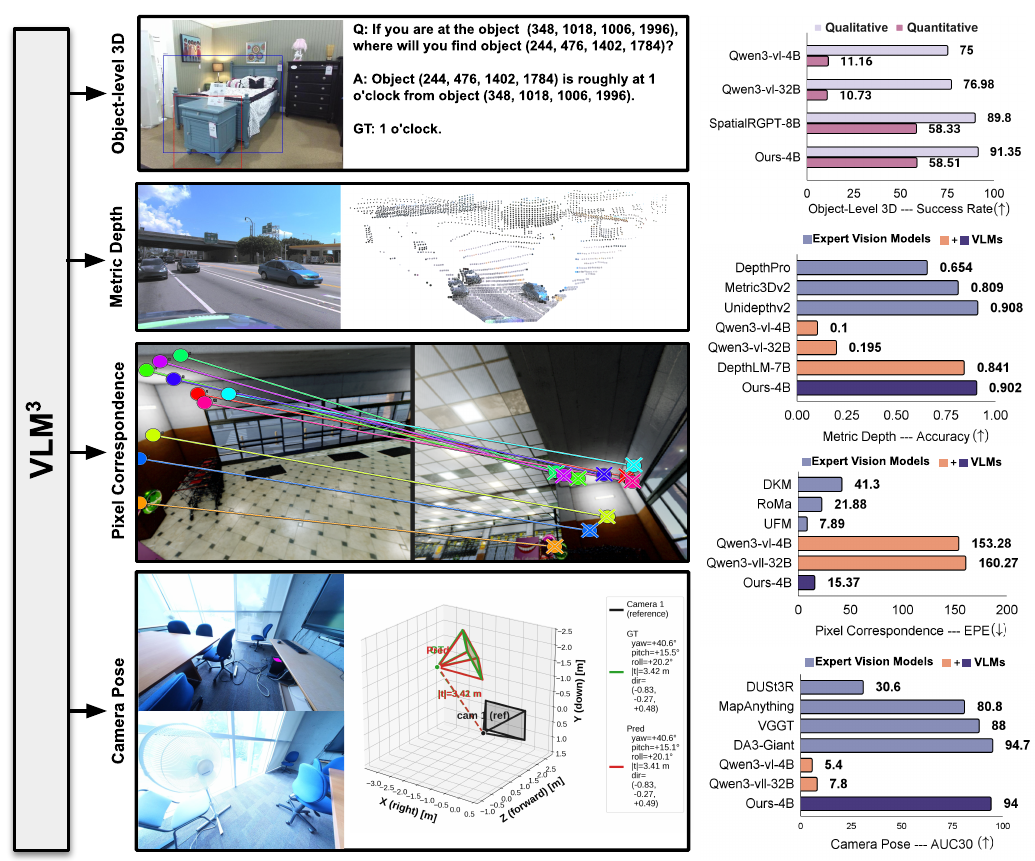}
    \caption{\textbf{We propose \methodname, a scalable method with the simplest design showing that VLMs are naitive 3D learners.} \methodname\ enables standard VLMs to learn diverse and fine-grained 3D tasks, matching expert vision models with complex task-specific design. For depth estimation, the numbers are averaged across: NuScenes, ETH3D, SUNRGBD, and iBims1, same as~\citep{cai2025depthlm}. Other numbers are from Table~\ref{tab:main_result} and \ref{tab:main_result_cv}. All visualization results are converted from text outputs (see Sec.~\ref{sec:method} for prompts). The bounding boxes in the object-level 3D example is only for visualization purposes. For pixel correspondence, the lines with dot ends are the prediction, the cross is the ground-truth (GT).}
    \label{fig:Teaser}
\end{figure}

Understanding 3D from 2D inputs lies at the core of visual intelligence. Vision Language Models (VLMs)~\citep{liu2023visual} allow a unified model to solve various vision tasks through prompting. Though effective in semantic understanding, existing VLMs still struggle with 3D understanding, especially for fine-grained tasks. As a result, expert vision models~\citep{metric3dv2, edstedt2024roma, lin2025depth} with complex task-specific design in data augmentations, architectures and losses are still the dominant approaches. 

Recently, DepthLM~\citep{cai2025depthlm} shows that standard VLMs can learn pixel-level depth estimation. Inspired by this observation, we ask in this work: \emph{``Can standard VLMs, without complex task-specific design, match expert vision models in diverse, fine-grained 3D understanding tasks beyond depth estimation?''} Through extensive study, we show that the answer is \emph{yes}!

Though prior works have explored 3D understanding VLMs, most of them either focus on coarse-grained object-level understanding~\citep{spatialvlm}, which cannot match the performance of expert vision models in fine-grained tasks, or still require task-specific design such as extra encoders/modules~\citep{spatialrgpt, zhang2026generalization}, which makes their training/model not compatible with standard VLMs.

This work explores not only the object-level tasks but also the fine-grained 3D tasks where expert vision models dominate. Our in-depth, large scale study shows that standard VLMs with surprisingly simple design, which neither change the architecture/losses nor add heavy data augmentations, are already effective 3D learners. Most task-specific designs are not necessary conditions for effective 3D learning. 
These not only include the complex designs in expert vision models, but also include many designs in 3D understanding VLMs such as extra encoders in object-level understanding~\citep{spatialrgpt} and visual prompting in pixel-level understanding~\citep{cai2025depthlm}. Interestingly, we even show that the regression loss, which is the foundational formulation of many 3D tasks, is also \emph{not} needed. Treating inputs and outputs all as text is sufficient to reach similar accuracy.

Based on these findings, we propose \emph{\methodname}, a scalable and simple framework that allows standard VLMs to learn diverse 3D tasks and match expert vision model accuracy. At the core of \methodname\ are: 1) Focal length unification through image resizing, which solves the camera ambiguity problem and enables mix-data training. 2) Text-based pixel/region reference with normalized ranges for both horizontal and vertical axes, which removes the need for visual prompting in previous works~\citep{cai2025depthlm} and makes \methodname\ simpler, much more efficient and scalable. 3) Data mixture and scaling, which turn out to be much more important than designing complex data augmentations, architectures and losses.

As shown in Fig.~\ref{fig:Teaser}, \methodname\ for the first time, enables standard VLMs to learn accurate 3D understanding across diverse and fine-grained tasks, including 1) object-level 3D understanding, 2) metric depth estimation, 3) pixel correspondence estimation, and 4) camera pose estimation. 
\begin{itemize}
    \item For \emph{object-level 3D understanding}, \methodname-4B improves over SpatialRGPT-8B~\citep{spatialrgpt} on SpatialRGPT-Bench while removing the need for extra encoders.
    \item For \emph{depth estimation}, \methodname-4B improves the accuracy of the previous best VLM DepthLM-7B~\citep{cai2025depthlm} from 0.84 to 0.9, matching the accuracy of UnidepthV2~\citep{unidepthv2}.
    \item For \emph{pixel correspondence}, \methodname\ reduces the EPE of the base VLM~\citep{bai2025qwen3} by 10x and  outperforms expert vision models such as DKM~\citep{edstedt2023dkm} and RoMa~\citep{edstedt2024roma}.
    \item For \emph{camera pose estimation}, \methodname\ improves the AUC30 of the base VLM from 5\% to 94\%, surpassing VGGT~\citep{wang2025vggt} and matching the accuracy of DA3-Giant~\citep{lin2025depth}. 
\end{itemize}
Our findings provide a new perspective on what is and is not necessary for 3D vision. We hope they can motivate simpler and better design of foundation models in the future.

\section{Related Work}\label{sec:related}

\textbf{Task-specific Design in Expert Vision Models.} Expert vision models rely on task-specific designs for different 3D tasks. A pre-trained vision encoder~\citep{weinzaepfel2022croco, oquab2023dinov2} is often applied, together with multiple decoders and task-specific routings. The decoders have varied architectures such as DPT~\citep{ranftl2021vision}, FPN~\citep{lin2017feature, unidepthv2}, Gaussian Process~\citep{edstedt2023dkm}, self-attention + linear layers~\citep{wang2025vggt} etc.. 

Monocular depth estimation often involves multiple decoders~\citep{unidepthv2} for depth, confidence and optionally camera ray maps. Pixel correspondence~\citep{edstedt2023dkm, edstedt2024roma, shen2024gim} often relies on multi-scale warping for accurate matching. For camera pose estimation, SOTA models~\citep{wang2025vggt, lin2025depth} often combine the supervision from multiple tasks such as depth, camera ray, point tracks, and poses to boost each other. Having multiple prediction heads also leads to multiple complex losses including MSE~\citep{unidepthv2}, L1~\citep{unidepthv2}, certainty~\citep{edstedt2023dkm}, regression by classification~\citep{edstedt2024roma} and variants of them such as clipped L2~\citep{edstedt2023dkm} etc.. The complexity does not lie solely in the type of losses, but on the number of losses, where balancing weights need to be tuned for different methods and tasks. Besides architectures and losses, heavy data augmentations are often important for expert vision model design. A typical example~\citep{unidepthv2, wang2025vggt} often includes both geometric augmentations such as random resizing, cropping, translation, and photometric augmentations such as brightness, gamma, saturation, hue shift etc.

In this work, we challenge the necessity of task-specific designs and provide new perspectives on what is really important for 3D learning in the context of generalist models. We show that standard VLMs without task-specific design mentioned above, are sufficient for effective 3D learning and match the performance of heavily designed expert models.

\textbf{VLMs for 3D understanding.} Many existing works have already explored 3D understanding with VLMs. \cite{spatialvlm} converts expert vision model predictions into text prompts for training and evaluation, which works for object-level understanding.~\cite{spatialrgpt} further advaces this direction by separating qualitative and quantitative problems and designing extra encoders for object reference without the need for using object names in text. This is helpful especially when multiple objects of the same type are presented in the input image, and removes the semantic names from the input query, which allows the evaluation to really show the 3D understanding capabilities at the object-level. Several other works further expand the task and input diversity~\citep{SpatialBot, fan2025vlm}. However, most of them focus only on object/scene-level understanding, and require extra architectures such as encoders and other modules. 

Recent works~\citep{multispatialMLLM, seedvl, cai2025depthlm} start to investigate fine-grained 3D understanding such as depth estimation. DepthLM~\citep{cai2025depthlm} shows for the first time that a standard VLM can learn pixel-level metric depth estimation with comparable accuracy to expert vision models. Inspired by this observation, this work further expands the depth and scale of the study, allowing us to design a simpler and more scalable method, which shows that standard VLMs can actually learn diverse object-level, pixel-level, single-view and multi-view tasks all without task-specific design.  

\section{Method}\label{sec:method}

\begin{figure}
    \centering
    \includegraphics[width=1.0\linewidth]{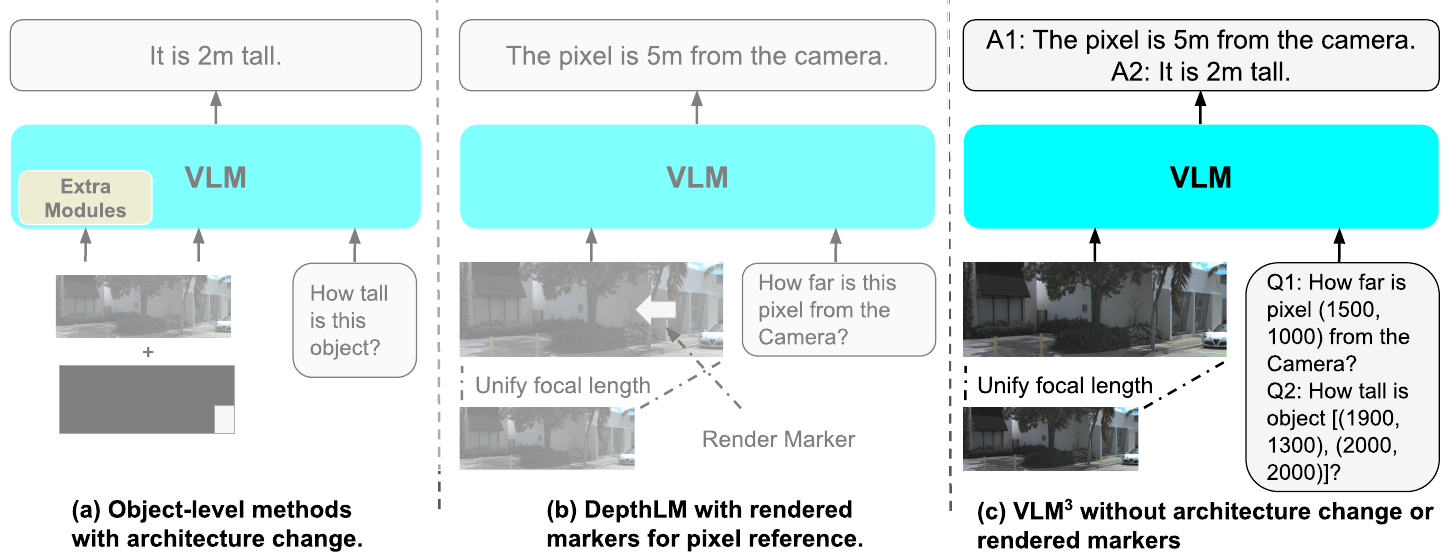}
    \caption{\textbf{\methodname\ overview}. Given the input images, \methodname\ first resizes them so that the focal length is 1000 pixels. This solves camera ambiguity without the need for adding extra VLM encoders/modules. To refer to an object or pixel, \methodname\ simply uses text with the pixel range normalized to [0, 2000) for both horizontal and vertical axes. This requires no architecture change~\citep{spatialrgpt} or marker rendering~\citep{cai2025depthlm}, and makes \methodname\ much more flexible and scalable. Standard VLM architectures and text-based training (SFT) are used to train the model.}
    \label{fig:pipeline}
\end{figure}

Fig.~\ref{fig:pipeline} shows the overview of \methodname. Given the input images, which can be one or multiple depending on the tasks, \methodname\ first resizes the images so that the focal length is 1000 pixels. As mentioned in DepthLM~\citep{cai2025depthlm}, this effectively addresses the camera ambiguity issue, and enables effective mixed-data training. Unlike DepthLM that requires rendered markers for pixel reference, \methodname\ directly refers to pixels/object regions in text by normalizing the pixel space to [0, 2000) in both horizontal and vertical axes. Standard text-based SFT is used to train the model on diverse tasks. Unless otherwise stated, we use Qwen3-vl-4B~\citep{bai2025qwen3} as our base VLM.

\subsection{Key Ingredients}\label{sec:method_key_ingredients}

\textbf{Images without camera intrinsics.} To maintain the simplicity and make \methodname\ fully compatible with standard VLM pre/post-training, we apply the same approach as DepthLM, i.e., unifying the focal length of input images, to solve the camera ambiguity. This approach requires no architecture change used in object-level spatial reasoning VLMs~\citep{spatialrgpt, zhang2026generalization}. 

However, there are cases where the images are from unknown sources and do not contain camera intrinsics information. In such cases, we simply apply pre-trained single image calibration models~\citep{tirado2025anycalib} to estimate the intrinsics so that we can still unify the focal length. For example, for the object-level 3D understanding experiment, we use this approach to obtain the intrinsics for both training and evaluation data, which works well in practice for in-the-wild images from the internet.

\textbf{Text-based pixel reference.} DepthLM uses visual prompting for pixel reference, which directly render markers on the input image. Though this approach works, it has limited scalability since training/inferencing on multiple pixels of the same image requires the same amount of input images with markers rendered on different places. As a result, DepthLM only trains the model on around 16M images + 2 pixels per image. On the other hand, relying on rendered markers makes it hard for tasks where the outputs also need pixel reference, e.g., pixel correspondence estimation~\citep{edstedt2023dkm}.

To address this issue, we conduct further analysis on pixel reference strategies, and find that though text-based pixel reference does not work with arbitrary prompt, it can be enabled with pixel space normalization. Specifically, \cite{cai2025depthlm} argue that VLMs do not understand text-based pixel reference. This argument is based on their experiment using the following prompt:  “Given this image of size (width = \texttt{w},
height = \texttt{h}), how far is the pixel at \texttt{(x, y)} from the camera?” Inspired by the VLM-based object detection methods~\citep{liu2025visual}, instead of telling the model the size of the image and the pixel location, we conduct further analysis where we use the prompt of "How far is the pixel at \texttt{(x, y)} from the camera? Both \texttt{x} and \texttt{y} are normalized to between [0, 2000)." As shown later in Sec.~\ref{sec:exp_analysis}, normalizing pixel space, surprisingly, allows text-based pixel reference to achieve similar accuracy as visual prompting. This shows that normalizing pixel space is an effective reference approach that works not only for coarse-grained object regions but also fine-grained pixel locations.

Text-based reference not only removes redundant image augmentations, but also makes \methodname\ much more efficient --- we can pack multiple questions for the same image(s) during training/inference without duplicating images for different questions. This allows us to use much less compute or enable a much larger training scale. For example, for depth estimation, instead of training on 1 labeled pixel per sample, we can now train on 10 labeled pixels per sample with negligible computation overhead. 

On the other hand, it also allows us to use the same simple method to handle diverse 3D tasks such as object-level 3D understanding where we can use text to refer to different objects, and pixel correspondence where we can treat both the query and output pixel locations as text.

\noindent\textbf{Data mixture and scaling.} A key insight of \methodname\ is that \emph{once camera ambiguity and pixel reference problems are solved, scaling up data is sufficient for standard VLMs to learn accurate 3D understanding}. Complex task-specific designs are not necessary conditions.

Unlike DepthLM where most of the datasets have uniform weights, Sec.~\ref{sec:exp_analysis} shows that data mixture becomes (almost) the most important thing when we scale up training. When using diverse training datasets with drastically different sizes, naively scaling up training without proper weighting often leads to saturated or even worse performance. This is because smaller or simpler datasets can be easily overfitted by VLMs with billions of parameters, which should be assigned smaller weights. In our experiments, weighting datasets based on their sizes is a reasonable baseline that works across tasks, though further tuning still has the potential to improve the performance significantly.

\subsection{Enable Diverse Tasks}\label{sec:method_diverse_tasks}

To verify the generality of \methodname, we choose 4 mainstream 3D understanding tasks with sufficient diversity, covering both single- and multi-view settings and requiring drastically different designs in previous models: 1) Metric depth estimation; 2) Object-level 3D understanding; 3) Pixel correspondence estimation; 4) Camera pose estimation. We introduce in this section how we enable each task. Further details are reported in Appendix~\ref{appdx:implementation}. 

\noindent\textbf{Metric depth estimation.} Following DepthLM, we formulate metric depth estimation as estimating the distance between query pixels to the camera. Most of our settings follow DepthLM except: 1) we use text-based pixel reference and pack 10 QAs corresponding to 10 labeled pixels of the same image for each training sample; 2) we add 10M internal images of outdoor street views to the original data mixture of DepthLM. This pushes the training data size from 16M to 26M. 3) To verify the importance of data mixture ratio, we conduct an in-depth analysis in Sec.~\ref{sec:exp_analysis}. As a result, we apply a non-uniform dataset weighting as reported in Appendix~\ref{appdx:implementation}. We train our model on 32M samples (320M labeled pixels). Unlike DepthLM that requires 128 H100 GPUs + 2 days to train a smaller model (3B) on $\frac{1}{10}$ of the labeled pixels, our training is done with 32 GPUs + 3 days.

\noindent\textbf{Object-level 3D understanding.} We train and evaluate our model on the same datasets as SpatialRGPT~\citep{spatialrgpt}, which includes both qualitative and quantitative questions. The training is done on 1M images, which requires 32 GPUs + 3 hours. Different from SpatialRGPT that requires extra encoders to encode object region masks, we simply use the bounding box coordinates \texttt{(xMin, yMin, xMax, yMax)} in text to refer to each object. The remaining prompts follow exactly the original format. We use a pretrained single image calibration model~\citep{tirado2025anycalib} to estimate the camera intrinsics for each image and enable focal length unification.

\noindent\textbf{Pixel correspondence estimation.} Pixel correspondence is a popular multi-view task~\citep{hartley2003multiple}. The goal is, for a query pixel in the left image, to find the corresponding pixel in the right image. For training, we use a mixture of datasets consisting of roughly 10M image pairs (see Appendix~\ref{appdx:implementation}). For simplicity we do not tune the data mixture and simply use the number of image pairs per dataset as the weighting, which works reasonably well in practice. For evaluation, we follow the metric (EPE) and datasets in UFM~\citep{zhang2025ufm}. The training is done on 80M samples + 10QA per sample, which requires 64 GPUs + 7 days. We simply use 5 randomly generated prompt templates from LLMs, and in practice the model is not sensitive to the prompt format. An example prompt we use is: ``Question: Given these two images, what pixel in the second image corresponds to pixel \texttt{(x1, y1)} in the first image? Report the answer as (x2, y2). Answer: The corresponding pixel is \texttt{(x, y)}." Note that pixel correspondence does not require understanding the metric scale of the world, and normalizing the focal length is not needed empirically, though it does not hurt the performance. 

\noindent\textbf{Camera pose estimation.} Camera pose estimation is another important multi-view task that has many applications. In our experiments, we use 2 images as inputs, and prompt the model to estimate the translation distance, translation direction and rotation direction. We use the same training data and mixture weights as in pixel correspondence estimation, and evaluate our models on ETH3D and ScanNet++ datasets using the same metric (AUC30$^\circ$) as in SOTA methods~\citep{lin2025depth, wang2025vggt}. We train our model on 10M samples, which can be done with 32 GPUs + 4 days. To enable text-based pose estimation, we represent 1) translation distance in meters, 2) translation direction as a unit 3D vector, and 3) rotation direction as yaw-pitch-roll numbers. Each pose component forms a unique question and we pack all questions into the same sample during training and evaluation. An example prompt is as below: 
\begin{itemize}
    \item Translation distance: "Question: Estimate the magnitude of the camera translation between the two viewpoints. Answer: Translation distance: \texttt{x} meters."
    \item Translation direction: "Question: Using the first camera as the reference frame, describe the displacement of the camera between the two views.  Use the first camera's local axes (X = right, Y = down, Z = forward) and give a qualitative direction together with the precise unit direction vector (x, y, z). Answer: The camera moves \texttt{[coarse direction, e.g., right, backward]}, unit vector \texttt{(x, y, z)}."
    \item Rotation direction: "Question: Treating the first image as the reference viewpoint, describe the camera's reorientation needed to reach the second image's viewpoint as yaw, pitch and roll about the first camera's local axes, applied intrinsically in the order yaw -> pitch -> roll.  Conventions: yaw is rotation about the vertical (down) axis, positive = turn right; pitch is about the lateral (right) axis after yaw, positive = look up; roll is about the optical (forward) axis after yaw and pitch, positive = bank to the right (image content rotates clockwise from the operator's POV). Answer: \texttt{Yaw=x}, \texttt{Pitch=y}, \texttt{Roll=z}."
\end{itemize}

Achieving SOTA accuracy in pose estimation in such a simple fashion is much more surprising to us compared to other tasks. As in traditional vision approaches, pose estimation is often done either by multi-step approaches~\citep{schonberger2016structure} (estimate pixel correspondences -> solve optimization problems), or learned with complex regression losses~\citep{wang2025vggt, lin2025depth} coupled with complementary tasks such as camera ray direction estimation, point track estimation, depth estimation to enable generalization. A standard VLM, without heavy tuning on the prompts, simply using next token prediction to output the text description of the poses, is arguably a completely new paradigm. This shows a clear signal that even the regression formulation, which is the foundation of most expert 3D vision models, is not a necessary condition for effective 3D learning. This holds true even for extremely challenging tasks requiring complex outputs. We believe this finding opens up a completely new perspective for 3D foundation models. 

\definecolor{lightgray}{gray}{0.7}

\begin{table*}[!htb]
\centering
    \caption{\textbf{Comparison with VLMs.} \methodname\ enables standard VLMs to master diverse single and multi-view 3D understanding tasks. For single view tasks, \methodname\ demonstrates SOTA performance in both pixel-level metric depth estimation and object-level spatial reasoning at both qualitative level and quantitative level. For metric depth estimation, \methodname\ improves the accuracy of DepthLM from 0.84 to 0.9 with a smaller model (4B vs 7B) and simpler method (no marker-based pixel reference). For object-level spatial reasoning, \methodname\ improves the accuracy of SpatialRGPT in both qualitative and quantitative understanding, while removing the need of architecture change such as extra encoders for region-masks. For multi-view tasks, \methodname\ effectively learns accurate pixel correspondence and camera pose estimation, improving the baseline models by a large margin. The best result is \textbf{bold faced}.}
\resizebox{\textwidth}{!}{
\renewcommand{\arraystretch}{1.3}%
\begin{tabular}{@{}lrrrrrrrrr}
    \hline
    \multirow{2}{*}{\textbf{Metric Depth} ($\delta_1 (\uparrow)$)}            & \multicolumn{3}{c}{\cellcolor{blue!20}\textit{Out}} & \multicolumn{1}{c}{\cellcolor{yellow!20}\textit{Out+In}} & \multicolumn{4}{c}{\cellcolor{red!20}\textit{In}} & \\
                            &  Argoverse2                          & DDAD                          & NuScenes                          & ETH3D                          & ScanNet++                          & sunRGBD                          & iBims1                          & NYUv2 & Average \\
    \hline
    \textsc{Qwen2.5-VL-72b~\citep{Qwen2.5-VL}}                       &  0.119    & 0.140    & 0.186                         &         0.220                 &          0.272                & 0.276                         & 0.212                         &      0.324 &  0.219  \\
    \textsc{Qwen3-VL-4b~\citep{bai2025qwen3}}                       &  0.004    & 0.073    & 0.070                        &         0.071                 &          0.147                & 0.176                         & 0.080                         &      0.158 &  0.101  \\
    \textsc{Qwen3-VL-32b}                       &  0.017    & 0.099    & 0.029                        &         0.167                 &          0.373                & 0.463                         & 0.122                         &      0.393 &  0.208  \\
    \textsc{Gemini-2.5-Pro~\citep{gemini}} & 0.280 & 0.252 & 0.365 & 0.328 & 0.380 & 0.270 & 0.466 & 0.394 & 0.342\\
    \textsc{GPT-5~\citep{achiam2023gpt}} & 0.218 & 0.302 & 0.382 & 0.313 & 0.428 & 0.471 & 0.307 & 0.540 & 0.370 \\
     \rowcolor{gray!20}\multicolumn{10}{c}{\textit{Spatial VLMs}} \\
     \textsc{SpaceLLaVA-13B~\citep{space-llava}}                       &  0.100    &  0.067   &   0.083                       &       0.090                   &         0.269                 &                   0.233       &           0.208              &      0.178  & 0.154    \\
     \textsc{SpatialRGPT-8B~\citep{spatialrgpt}}                       &   0.055   &  0.046   &      0.100                    &         0.220                 &             0.346             &                  0.369        &            0.240             &      0.265  & 0.205   \\    
     \rowcolor{gray!20}\multicolumn{10}{c}{\textit{VLMs Trained on Metric Depth Estimation}} \\
    \textsc{Seed1.5-VL~\citep{seedvl}} & 0.040 & 0.074 & 0.028 & 0.309 & 0.593 & 0.689 & 0.627 & 0.841 & 0.400\\
    \textsc{DepthLM-3b~\citep{cai2025depthlm}}                       & 0.808     & 0.724    & 0.870                         & 0.745                         & 0.838                         & 0.850                         & 0.890                         & 0.868          & 0.824  \\
    \textsc{DepthLM-7b}                       & 0.833     & 0.747    & 0.865                         & 0.718                         & 0.850                         & 0.859                         & 0.920                         & 0.915  & 0.838               \\
    \rowcolor{blue!10}\textsc{\textbf{Ours-4b}}                       & \textbf{0.896}     & \textbf{0.818}    & \textbf{0.970}                         & \textbf{0.810}                         & \textbf{0.976}                         & \textbf{0.867}                         & \textbf{0.960}                         & \textbf{0.935}     & \textbf{0.904}           \\
    \hline
\end{tabular}
}

\vspace{0.5em}

\resizebox{\textwidth}{!}{
\renewcommand{\arraystretch}{1.3}%
\begin{tabular}{@{}lrrrrrrr}
    \hline
    \textbf{Object-level 3D Understanding}            
                            & \multicolumn{7}{c}{\cellcolor{gray!20}\textit{SpatialRGPT-Bench (Qualitiative)}} \\ 
                            (Acc $(\uparrow)$) & Below/Above &	Left/Right & Big/Small	& Tall/Short &	Wide/Thin	& Behind/Front	& Overall           \\
    \hline
    \textsc{Qwen3-VL-4b}        &  63.33	& 85.71	& 85.85	& 70.54 & 83.65	& 60.91	& 75.00                 \\
    \textsc{Qwen3-VL-32b}        & 72.50	& 83.81	& 82.08	& 68.75 & 87.5	& 67.27	& 76.98                 \\
    \textsc{SpatialRGPT-8b}        &  \textbf{99.17}	& 99.04	& 79.24	& 89.28	& 83.65	& \textbf{87.27}	& 89.80    \\
    \rowcolor{blue!10}
    \textbf{\textsc{Ours-4b}}               &  97.84	& \textbf{99.37}	& \textbf{90.78}	& \textbf{90.82}	& \textbf{92.86}	& 75.54	& \textbf{91.35} \\
    \hline
\end{tabular}
}

\resizebox{\textwidth}{!}{
\renewcommand{\arraystretch}{1.3}%
\begin{tabular}{lrrrrrrr}
    \hline
    \textbf{Object-level 3D Understanding}      & \multicolumn{7}{c}{\cellcolor{gray!20}\textit{SpatialRGPT-Bench (Quantitative)}} \\
    (Acc $(\uparrow)$ / AbsRel $(\downarrow)$)     &    Direct Distance &	Horizontal Distance	& Vertical Distance	& Width	& Height	& Direction	& Overall \\
    \hline
    \textsc{Qwen3-VL-4b}       &   18.24 / 51.94  &	18.03 / 80.64 &	14.15 / 160.62	& 3.76 / 552.54	& 12.78 / 874.07	& 0.0 / 180.0°	& 11.16 / 343.96 \\
    \textsc{Qwen3-VL-32b}       &   22.30 / 97.14  &	17.21 / 67.60 &	22.64 / 44.93	& 0.00 / 582.34	& 2.26 / 647.74	& 0.0 / 180.0°	& 10.73 / 287.95 \\
    \textsc{SpatialRGPT-8b}    &  \textbf{35.1 / 0.35}	& \textbf{59.0 / 0.27}	& 53.8 / 0.27	& \textbf{51.9 / 0.31}	& 54.9 / 0.63 &	95.3 / 17.1°	& 58.33 / 0.37 \\
    \rowcolor{blue!10}
    \textbf{\textsc{Ours-4b}}  & 34.09 / 0.37	& 53.38 / 0.29	& \textbf{58.41 / 0.27} &	44.11 / 0.39	& \textbf{65.64 / 0.41}	& \textbf{95.42 / 10.5°}	& \textbf{58.51 / 0.35} \\
    \hline
\end{tabular}
}

\vspace{0.5em}

\resizebox{\textwidth}{!}{
\renewcommand{\arraystretch}{1.3}%
\begin{tabular}{@{}lrrrr}
    \hline
    \textbf{Pixel Correspondence} (EPE$(\downarrow)$)
                            &  ETH3D                          & DTU                          & TA-WB                         & Average \\
    \hline
    \textsc{Qwen3-VL-4b}                       &  179.80   &  89.64   &  190.40                       &   153.28   \\
    \textsc{Qwen3-VL-32b}                       &  196.53   &  88.76   &    195.53                     &   160.27   \\
    \hline
    \rowcolor{blue!10}\textsc{\textbf{Ours-4b}}                       & \textbf{15.18}     & \textbf{10.71}    & \textbf{20.21}                         & \textbf{15.37}  \\
    \hline
\end{tabular}

\renewcommand{\arraystretch}{1.3}%
\begin{tabular}{@{}lrrr}
    \hline
    \textbf{Camera Pose} (AUC@30\degree $(\uparrow)$)           
                  & ETH3D          &  ScanNet++       & Average      \\
    \hline
    \textsc{Qwen3-VL-4b}        &       10.0        &  0.7 & 5.4 \\
    \textsc{Qwen3-VL-32b}       &         11.7       &  3.9 & 7.8 \\
    \hline
    \rowcolor{blue!10}\textsc{\textbf{Ours-4b}}           &     \textbf{93.3}       & \textbf{94.7} & \textbf{94.0} \\
    \hline
\end{tabular}

}

\label{tab:main_result}
\vspace{-3mm}
\end{table*}

\definecolor{lightgray}{gray}{0.7}

\begin{table*}[t]
\centering
    \caption{\textbf{Comparison with expert vision models.} Without heavy data-augmentation, change of standard VLM architecture and training losses, \methodname\ achieves comparable performance to expert vision models. In \emph{metric depth estimation}, \methodname\ matches the accuracy of MoGe-2 and UnidepthV2. In \emph{pixel correspondence}, it has lower EPE than DKM and RoMa. In \emph{camera pose estimation}, it outperforms VGGT and matches the DA3-Giant accuracy. The best results are \textbf{bold faced}.}
\resizebox{0.7\textwidth}{!}{
\renewcommand{\arraystretch}{1.3}%
\begin{tabular}{@{}lrrrrr}
    \hline
    \multirow{2}{*}{\textbf{Metric Depth} ($\delta_1 (\uparrow)$)}            & \multicolumn{2}{c}{\cellcolor{blue!20}\textit{Out}} & \multicolumn{1}{c}{\cellcolor{yellow!20}\textit{Out+In}} & \multicolumn{2}{c}{\cellcolor{red!20}\textit{In}} \\
                                                      & DDAD                          & Nuscenes                          & ETH3D                                                   & sunRGBD                          & ibims1                           \\
    
    \hline
    \textsc{ZoeDepth~\citep{zoedepth}}                   & 0.272    & 0.283                         & 0.350                                                & 0.867                         & 0.580                               \\
    \textsc{DepthAnything~\citep{yang2024depth}}                     & -    & 0.354                         & 0.093                                             & 0.850                         & 0.714                                        \\
    \textsc{DepthAnythingV2~\citep{yang2024depth}}                     & -    & 0.171                         & 0.363                                                 & 0.724                         & -                      \\
    \textsc{Metric3D~\citep{metric3d}}                     & -    & 0.723                         & 0.456                                            & 0.154                         & 0.797                                           \\
    \textsc{Unidepth~\citep{unidepth}}                     & 0.858                         & 0.846   & 0.185                                              & 0.943                         & 0.157                                         \\
    \textsc{Depth Pro~\citep{depthpro}}                    & 0.299    & 0.566                         & 0.397                                             & 0.831                         & 0.823                                            \\
    \textsc{Metric3Dv2~\citep{metric3dv2}}                    & -    & 0.841                         & 0.900                                                 & 0.812                         & 0.684                                  \\
    \textsc{UnidepthV2~\citep{unidepthv2}}                     & \textbf{0.882}    & 0.870                         & 0.852                                                 & \textbf{0.964}                         & 0.945                                           \\
    \textsc{MoGe-2~\citep{moge2}}                     & 0.856    & -                         & \textbf{0.908}                                                 & -                         & 0.924                                           \\
        \rowcolor{blue!10}\textsc{\textbf{Ours-4b}}                           & 0.818    & \textbf{0.970}                         & 0.810                                                & 0.867                         & \textbf{0.960}                                      \\
    \hline
\end{tabular}
}

\resizebox{\textwidth}{!}{
\renewcommand{\arraystretch}{1.3}%
\begin{tabular}{@{}lrrrr}
    \hline
    \textbf{Pixel Correspondence} (EPE$(\downarrow)$)
                            &  ETH3D                          & DTU                          & TA-WB                         & Average \\
    \hline
    \textsc{DKM~\citep{edstedt2023dkm}}                       &  30.83   &  30.15   &  61.80                       &   41.30   \\
    \textsc{RoMa~\citep{edstedt2024roma}}                       &  10.01   &  11.84   &     43.79                    &   21.88   \\
    \textsc{UFM~\citep{zhang2025ufm}}                       &  \textbf{3.83}   &  \textbf{7.28}   &       \textbf{12.56}                  &   \textbf{7.89}   \\
    \rowcolor{blue!10}\textsc{\textbf{Ours-4b}}                       & 15.18     & 10.71    & 20.21                         & 15.37  \\
    \hline
\end{tabular}

\renewcommand{\arraystretch}{1.3}%
\begin{tabular}{@{}lrrr}
    \hline
    \textbf{Camera Pose} (AUC@30\degree $(\uparrow)$)           
                          & ETH3D  &  ScanNet++   &   Average      \\
    \hline
    \textsc{DUSt3R~\citep{wang2024dust3r}}               &   27.3     &  33.9  & 30.6 \\
    \textsc{MapAnything~\citep{keetha2025mapanything}}           &    77.4        &  84.1 & 80.8 \\
    \textsc{VGGT~\citep{wang2025vggt}}                  &   80.8   &  95.1 & 88.0 \\
    \textsc{DA3-Giant~\citep{lin2025depth}}              &    91.2     &  \textbf{98.1} & \textbf{94.7} \\
    \rowcolor{blue!10}\textsc{\textbf{Ours-4b}}            &    \textbf{93.3}     & 94.7 & 94.0 \\
    \hline
\end{tabular}
}

\label{tab:main_result_cv}
\vspace{-3mm}
\end{table*}

\section{Experiment}\label{sec:exp}

\noindent\textbf{Evaluation.} 1) For \emph{depth estimation}, we follow DepthLM~\citep{cai2025depthlm}, which uses 9 datasets to compare with VLMs and 5 datasets to compare with expert vision models. We use $\delta_1$ as the metric to measure the prediction accuracy. 2) For \emph{object-level 3D understanding}, we follow the same evaluation data and metrics as in SpatialRGPT~\citep{spatialrgpt}. 3) For \emph{pixel correspondence}, we follow the evaluation dataset and the EPE metric of UFM~\citep{zhang2025ufm}. EPE represents the error in terms of the number of pixels of the prediction in the target image domain. To make the evaluation directly comparable with expert vision models, we rescale the range of EPE to be the same as in UFM. We evaluate on 8192 samples per dataset similar to the depth estimation setting, and re-evaluate the expert vision models on the same data. 4) For \emph{pose estimation}, we report the AUC30$^\circ$ metric~\citep{lin2025depth} on ETH3D and ScanNet++ dataset. 

\subsection{Main Results}\label{sec:exp_main_result}

We report the comparisons with VLMs and expert vision models respectively in Table~\ref{tab:main_result} and~\ref{tab:main_result_cv}. 

\noindent\textbf{Metric depth estimation.} \methodname\ significantly surpasses SOTA VLMs, improving the accuracy of DepthLM-7B across \emph{all} datasets, pushing the average accuracy from 0.84 to 0.9, with about half the model size. Comparing to expert vision models such as MoGe-2 and UnidepthV2, \methodname\ also shows competitive accuracy, achieving new SOTA on NuScenes and iBims1 datasets.

\noindent\textbf{Object-level 3D understanding.} \methodname\ surpasses the accuracy of SpatialRGPT-8B for both qualitative and quantitative tasks. Meanwhile, unlike SpatialRGPT-8B that requires extra encoders for object reference, \methodname\ maintains the architecture of the base model and is much smaller.

\noindent\textbf{Pixel correspondence.} \methodname\ reduces the EPE of baseline VLMs by an order of magnitude. Comparing to expert vision models, \methodname\ shows competitive performance and has lower EPE than DKM and RoMa. Though it falls behind UFM, we believe with further scaling and more careful data mixture tuning, its performance can be further improved in the future.

\noindent\textbf{Camera Pose.} \methodname\ improved the accuracies of baseline VLMs by a large margin. Comparing to expert vision models, \methodname\ achieves similar accuracy to DA3-Giant (94.0 vs 94.7), surpassing mainstream models such as VGGT, MapAnything etc.

\begin{figure}[!htb]
    \centering
    \includegraphics[width=1.0\linewidth]{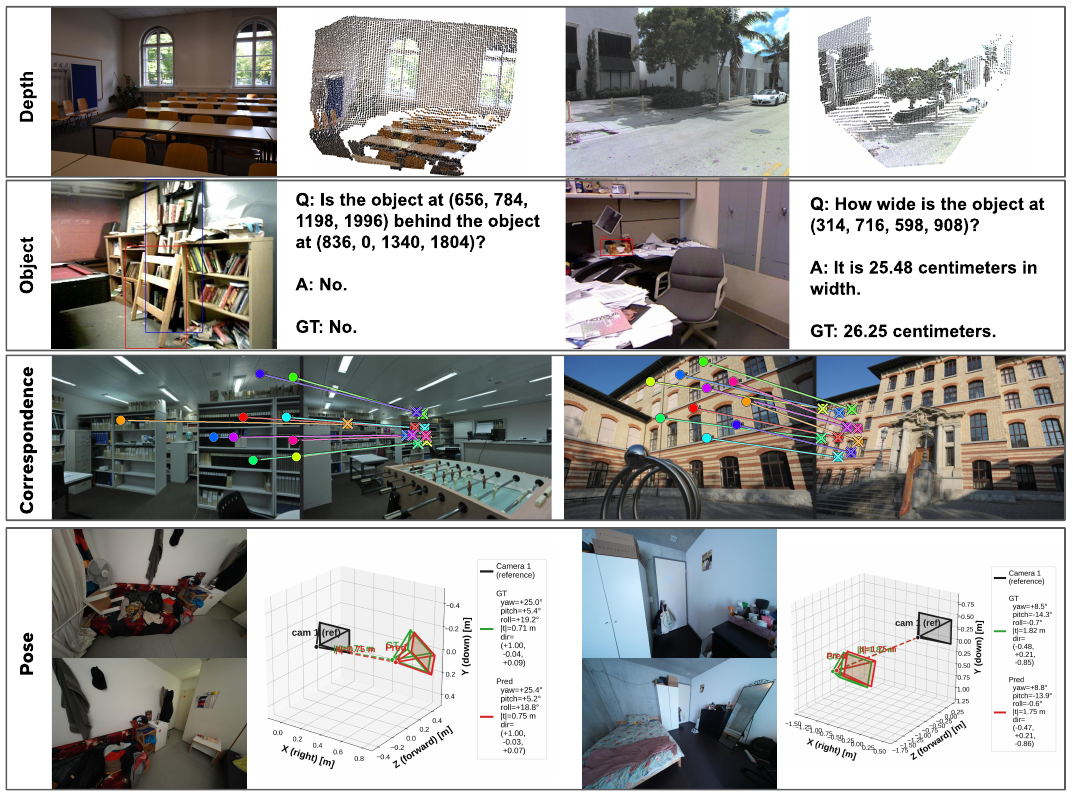}
    \caption{\textbf{Visualizations}. Consistent with the quantitative results, \methodname\ works across diverse tasks with both single and multi-view inputs, and both indoor and outdoor scenes. The bounding boxes in object-level 3D understanding are rendered only for visualization purposes. The model sees only the raw images during both training and evaluation. For pixel correspondence, we render both the predicted correspondences as lines with dotted ends. And the GT as the cross. For pose estimation, we convert the output predictions from text to rendered camera poses for more interpretable visualization, the actual predicted and GT numbers can be found in the rendered figure as well.}
    \label{fig:visualization}
\end{figure}

\noindent\textbf{Visualization.} Fig.~\ref{fig:visualization} visualizes the outputs of \methodname. Following DepthLM, we prompt \methodname\ on multiple pixels of the image to produce a dense point cloud. We find empirically that the performance remains similar no matter whether we prompt the model to predict each pixel independently or simply pack multiple queries for the same image. Note that the independent query strategy can also be implemented into an efficient method where all queries share the same pre-computed visual token set and text prefix part that remain the same across samples.

For \emph{depth estimation}, \methodname\ produces high-quality point clouds for both indoor and outdoor scenes. Similar to the observation of DepthLM, learning depth estimation via simple text supervision avoids flying points between 2 objects, which are often seen in expert vision models. We conjecture that this is due to the minimal inductive bias in \methodname\ and DepthLM, which is heavy in the task-specific design of expert vision models. For \emph{object-level 3D understanding}, \methodname\ can learn both the object-level spatial relationship such as front/behind, and metric-scale object properties. For \emph{pixel correspondence}, \methodname\ can predict reliable correspondences for both indoor and outdoor scenes. For \emph{camera pose estimation}, \methodname\ not only returns accurate rotation and translation directions, but also can predict the metric scale translation distance.

\subsection{Analysis}\label{sec:exp_analysis}

\definecolor{lightgray}{gray}{0.7}

\begin{table*}[t]
\centering
    \caption{\textbf{Analysis.} Left: text-based pixel reference performs similarly well as visual prompting. Mid: weighting datasets based on their sizes is a good baseline for large scale training, while further tuning still have large improvement room. Right: small models are sufficient to reach SOTA.}
\resizebox{\textwidth}{!}{
\renewcommand{\arraystretch}{1.3}%
\begin{tabular}{@{}lr}
    \hline
    \textbf{Pixel reference (8M samples + 1QA)}
                            &  $\delta_1$ ($\uparrow$) \\
    \hline
    \textsc{Visual Prompting}                       &  0.849   \\
    \textsc{Text-based}                       &   \textbf{0.853}   \\
    \hline
\end{tabular}

\renewcommand{\arraystretch}{1.3}%
\begin{tabular}{@{}lr}
    \hline
    \textbf{Data mixture (32M samples + 10 QA)}
                            &  $\delta_1$ ($\uparrow$) \\
    \hline
    \textsc{Uniform Weight}                       &  0.842   \\
    \textsc{Dataset-size based Weight}                       &  0.884   \\
    \textsc{\methodname\ Weight}                       &  \textbf{0.904}   \\
    \hline
\end{tabular}

\renewcommand{\arraystretch}{1.3}%
\begin{tabular}{@{}lr}
    \hline
    \textbf{Model and Dataset Size}
                            &  $\delta_1$ ($\uparrow$) \\
    \hline
    \textsc{32B (32M samples + 10 QA)}                       &  0.873   \\
    \textsc{8B (32M samples + 10 QA)}                       &   0.880   \\
    \textsc{4B (64M samples + 10 QA)}                       &   0.880   \\
    \textsc{4B (32M samples + 10 QA)}                       &   \textbf{0.904}   \\
    \hline
    
\end{tabular}
}
\label{tab:analysis}
\vspace{-3mm}
\end{table*}

\noindent\textbf{Besides simpler and more scalable, text-based pixel reference works similarly well as visual prompting.} As mentioned in Sec.~\ref{sec:method_key_ingredients}, text-based pixel reference is the key for \methodname\ to be simple, efficient and scalable. To verify its effectiveness compared to visual prompting, we conduct analysis experiments by training 2 models with different pixel reference approaches on 8M images + 1 QA per image. As shown in Table~\ref{tab:analysis}, given the same data and model, visual prompting and text-based pixel reference achieve similar accuracy. 

\noindent\textbf{Data mixtures are vital.} As mentioned in Sec.~\ref{sec:method_key_ingredients}, data mixture plays a vital role in 3D learning where mixed datasets with varied sizes are typically used. To show how much improvement we can get from careful data mixture, we compare in Table~\ref{tab:analysis} the performance of the models using varied data mixtures. For \emph{uniform weight}, we follow the same weighting as in DepthLM and set the weight of the newly added dataset to the same as other datasets (except for Matterport3d which has 0.1 weight and is the default of DepthLM). For \emph{dataset-size based weight}, we simply weight the datasets based on the number of images in them. For \emph{\methodname\ weight}, we follow Appendix~\ref{appdx:implementation} and further reduce the weighting for some datasets that were small and can be easily overfitted on. 

As shown in Table~\ref{tab:analysis}, \emph{uniform weight} with 32M samples + 10QA actually performs worse than simply training on 8M+1QA samples as in \emph{text-based} (left table), which uses $\frac{1}{40}$ labels. Therefore, naive scaling without careful data re-weighting cannot improve the model performance. Meanwhile, \emph{dataset-size based weight} is a good baseline for scaling up training, which improves the accuracy from 0.84 to 0.88. \emph{\methodname\ weight} further improves the accuracy to 0.9.

\noindent\textbf{Small VLMs are sufficient.} We use a 4B model in \methodname, one natural question is: \emph{can larger models further improve the performance?} As shown in Table~\ref{tab:analysis}, increasing the model size actually reduced the depth estimation accuracy. We conjecture that this is because our current dataset size is still not big enough for larger models, which can easily overfit. To further verify that this is the case, we scale up our training on the 4B model to 64M samples + 10QA, which we also observe a reduced accuracy, indicating that even for 4B model, it overfits the datasets at a slightly bigger scale. 

This result shows that at the level of 26M images, scaling data is still much more important than scaling the model size. Even 4B models can easily overfit to 26M images with slightly longer training. On the other hand, we can still reach SOTA accuracy with small 4B models.

\section{Conclusion}\label{sec:conclusion}

We propose \methodname, a scalable method with minimal design, proving for the first time that VLMs are native 3D learners. Specifically, with standard architecture and text-based SFT, VLMs can learn accurate 3D understanding across highly diverse tasks and match expert vision models consistently. The simplicity, flexibility and scalability of \methodname\ open up a new way to build generalist 3D foundation models. We also hope our findings can motivate the community to re-think what is and is not necessary for effective 3D learning. 

\clearpage
\newpage
\bibliographystyle{assets/plainnat}
\bibliography{paper}

\clearpage
\newpage
\beginappendix

\section{Further Implementation Details}\label{appdx:implementation}

\definecolor{lightgray}{gray}{0.7}

\begin{table*}[!htb]
\centering
    \caption{\centering\textbf{Hyper-parameters.}}
\resizebox{\textwidth}{!}{
\renewcommand{\arraystretch}{1.3}%
\begin{tabular}{@{}lrrrr}
    \hline
    \textbf{Task}
                            &  Depth Estimation & Object-level 3D & Pixel correspondence & Camera pose estimation \\
    \hline
    \textsc{Learning rate}                     & 5.5e-5 & 3.5e-4 & 2e-5 & 5e-5  \\
    \textsc{Batch size}                 & 1344 &  640 &  2816 & 448 \\
    \textsc{Number of samples}                 & 32M (10 pixels each) &  1M & 80M (10 pixels each) &  10M  \\
    \hline
\end{tabular}
}
\label{tab:implementation}
\end{table*}

\begin{table*}[!htb]
\centering
    \caption{\centering\textbf{Training data statistics.}}
\resizebox{\textwidth}{!}{
\renewcommand{\arraystretch}{1.3}%
\begin{tabular}{@{}lrr}
    \hline
    \textbf{Depth Estimation Datasets}
                            &  Number of images used & mixture weights \\
    \hline
    \textsc{Argoverse2~\citep{argoverse}}              & 1M &  0.21  \\
    \textsc{Waymo~\citep{waymo}}                 & 700K &   0.04  \\
    \textsc{Nuscenes~\citep{nuscenes}}     & 200K &   0.01  \\
    \textsc{ScanNet++~\citep{scannet++}}     & 1M &   0.01  \\
    \textsc{Taskonomy~\citep{taskonomy}}     & 4M &   0.51  \\
    \textsc{HM3d~\citep{hm3d}}     & 9M &   0.78  \\
    \textsc{Matterport3d~\citep{matterport3d}}     & 190K &   0.006  \\
    \textsc{Internal data with street view scenes}     & 10M &   1.0  \\
    \hline
\end{tabular}
}

\resizebox{\textwidth}{!}{
\renewcommand{\arraystretch}{1.3}%
\begin{tabular}{@{}lr}
    \hline
    \textbf{Pixel correspondence and camera pose datasets}
                            &  Number of image pairs used (same for mixture weights) \\
    \hline
    \textsc{BlendedMVS~\citep{yao2020blendedmvs}}              & 450K   \\
    \textsc{dynamicreplica~\citep{karaev2023dynamicstereo}}                 & 1M  \\
    \textsc{sailvos3d~\citep{hu2021sail}}     & 350K  \\
    \textsc{ScanNet++~\citep{scannet++}}     & 1M  \\
    \textsc{MPSD~\citep{antequera2020mapillary}}     & 13K \\
    \textsc{RealEstate-10K~\citep{zhou2018stereo}}     & 880K \\
    \textsc{DL3dv-10k~\citep{ling2024dl3dv}}     & 2.6M  \\
    \textsc{MegaDepth~\citep{matterport3d}}     & 190K \\
    \textsc{Aria Synthetic Environment~\citep{avetisyan2024scenescript}}     & 2M \\
    \textsc{GTA-SFM~\citep{wang2020flow}}     & 90K \\
    \textsc{Tartanairv2~\citep{wang2020tartanair}}     & 850K \\
    \textsc{UnrealStereo4K~\citep{tosi2021smd}}     & 270K \\
    \textsc{MVS Synth~\citep{huang2018deepmvs}}     & 190K \\
    \textsc{Spring~\citep{mehl2023spring}}     & 20K \\
    \hline
    \textsc{total}     & 9.9M \\
    \hline
\end{tabular}
}
\label{tab:datasets}
\end{table*}

Table~\ref{tab:implementation} and~\ref{tab:datasets} report hyperparameters and training data statistics for each task. In all our model training, we use a cosine learning rate schedule with linear warmup, with warmup ratio set to 0.1. We use AdamW optimizer with the default settings in the Transformers library. We use FSDP hybrid shard, gradient clipping of 0.02, gradient checkpointing, bfloat16, and Flash Attention 2.

To construct the training data for multiview tasks like pixel correspondence and camera pose estimation, we follow similar approaches to existing works~\citep{keetha2025mapanything} to randomly sample image pairs with $>25\%$ covisibility. Similar to previous works~\citep{cai2025depthlm, lin2025depth}, we hold out 30 scenes from ScanNet++ to ensure the evaluation data come from unseen scenes.

\end{document}